\documentclass[conference]{IEEEtran}

\IEEEoverridecommandlockouts
\usepackage{cite}
\usepackage{amsmath,amssymb,amsfonts}
\usepackage{algorithmic}
\usepackage{graphicx}
\usepackage{textcomp}
\usepackage{xcolor}
\usepackage{array}
\usepackage{caption}
\usepackage{subcaption}
\def\BibTeX{{\rm B\kern-.05em{\sc i\kern-.025em b}\kern-.08em
    T\kern-.1667em\lower.7ex\hbox{E}\kern-.125emX}}

\begin{document}

\title{An Android Robot Head as Embodied Conversational Agent}

\author{\IEEEauthorblockN{1\textsuperscript{st} Marcel Heisler}
\IEEEauthorblockA{
\textit{Hochschule der Medien}\\
Stuttgart, Germany \\
heisler@hdm-stuttgart.de}
\and
\IEEEauthorblockN{2\textsuperscript{nd} Christian Becker-Asano}
\IEEEauthorblockA{
\textit{Hochschule der Medien}\\
Stuttgart, Germany \\
becker-asano@hdm-stuttgart.de}
}

\maketitle

\begin{abstract}
This paper describes, how current Machine Learning (ML) techniques combined with simple rule-based animation routines make an android robot head an embodied conversational agent with ChatGPT as its core component. The android robot head is described, technical details are given of how lip-sync animation is being achieved, and general software design decisions are presented. A public presentation of the system revealed improvement opportunities that are reported and that lead our iterative implementation approach.
\end{abstract}

\begin{IEEEkeywords}
humanoid robotics, machine learning, software development, conversational agents
\end{IEEEkeywords}

\section{Introduction}
The advancements in research on android robots open up more and more application opportunities. For example,~android \textit{ERICA} \cite{kawahara_intelligent_2021} was already tested for attentive listening, job interview practicing, speed date practicing and as a lab guide. \textit{ERICA} was proposed to be employed in other social interaction tasks, e.g., as an attendant or a receptionist. Recent research suggests that android robots might be useful as interaction partners for community-dwelling older adults with little company \cite{carros_not_2022}. Additionally android robots might be useful tools in other research areas, too, e.g., in psychological studies regarding emotional interactions \cite{sato_android_2022}.

While such use cases provide promising perspectives for android robotics, they are often carried out using scripted actions or as Wizard of Oz studies. To actually employ such robots in real world scenarios, however, they need to act autonomously. As a first step in this direction, this paper describes how an android robot head is programmed to converse autonomously.


In contrast to other research publications describing android robot software architectures \cite{glas_erica_2016, minato_study_2022}, here a less complete, but much simpler approach is presented and described along with an iterative development approach. Manually defined animations are the basis for an implementation that heavily relies on machine learning (ML) models to achieve an embodied conversational agent \cite{cassell2001embodied} that is represented by an expressive android head.

First, in Section~\ref{background} some background information about the robot hardware and ML models in use is provided. Next, the current state of the implementation as well as the development process to reach this is described in Section~\ref{implementation}. Finally, the current state is briefly evaluated and next steps to implement are assessed in Section~\ref{evaluation}.

\section{Background}\label{background}


\subsection{Android Robot Head}
\begin{figure}[htbp]
\centerline{\includegraphics[width=0.45\textwidth]{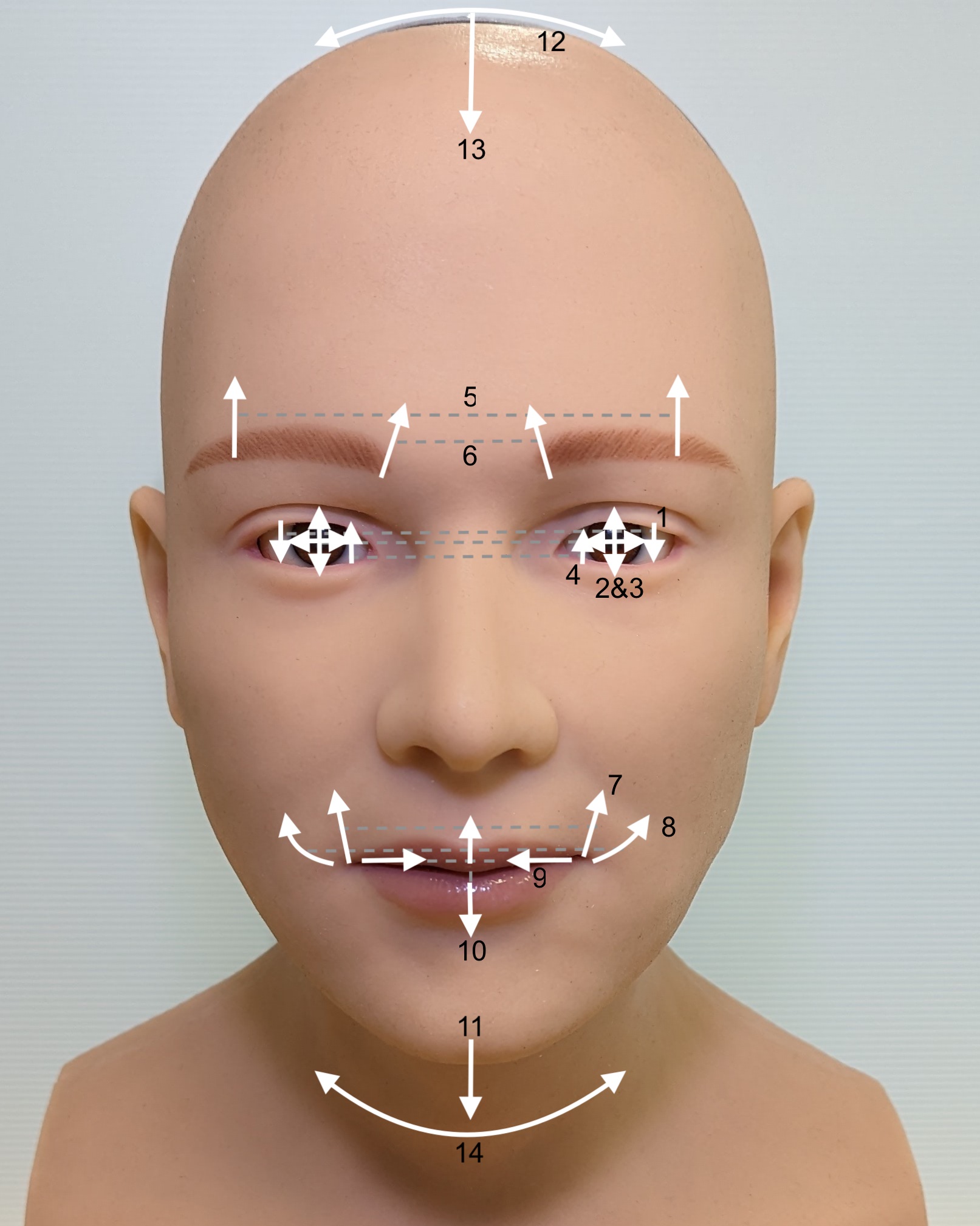}}
\caption{Actuators of the android robot head. Dotted lines indicate symmetric movements by a single actuator.}
\label{actuators}
\end{figure}

The android robot head was manufactured in Japan by the company A-Lab\footnote{https://www.a-lab-japan.co.jp/en.html}, cf.~\cite{heisler_making_2023} for further details. Its 14 pneumatic actuators shown in Fig.~\ref{actuators} enable it to display various facial expressions, e.g., to mimic emotions or lip-sync speech signals. The actuators are controllable by sending 14 integer values, each ranging from 0-255, via a RS-485 connection. The robot head does not provide any built in sensors to perceive its environment like cameras or microphones, nor a built in speaker. For the application described here external speakers and microphones are being used. The following actuators are available:

\begin{enumerate}
    \item upper eyelid down
    \item eyeball left right
    \item eyeball up down
    \item lower eyelid up
    \item eyebrow up
    \item eyebrow shrink
    \item mouth corner up
    \item mouth corner back
    \item lip shrink
    \item lips open
    \item jaw down
    \item lean head
    \item nod
    \item tilt head
\end{enumerate}

\subsection{ML Models}\label{ml_models}
Our implementation of an embodied conversational agent solves the following tasks using ML models: Automatic Speech Recognition (ASR), speech synthesis or Text To Speech (TTS), textual conversation or dialogue (chat) and automatic lip-sync. Related ML approaches for each of these tasks are described in the following paragraphs:

\paragraph{ASR} The current state of the art openly accessible ML model for ASR is \textit{Whisper}\cite{radford_robust_2022}. It consists of an off-the-shelf encoder-decoder Transformer architecture, which is known to scale well with increasing amounts of training data. The amounts of data used for training are also the main reason for Whisper to outperform previous models. 
Another notable novelty is that Whisper is trained on multiple tasks and multiple languages. Thus, a single Whisper model is not only capable of transcribing speech in one language but also to do tasks like voice detection, language identification, speaker diarization and translation from different languages to English. There are models in different sizes publicly available providing the common trade-off between more accurate results but slower inference times and stronger hardware requirements with increasing model sizes.

\paragraph{TTS} Over the last years ML was also adopted for speech synthesis. A development can be observed from multi-stage approaches, where first a model predicts acoustic features, e.g. mel-spectrograms, from linguistic features, e.g. raw text or phonemes, and in the next stage a vocoder model generates waveform from the acoustic features, towards end-to-end models, that generate waveform directly from linguistic features \cite{tan_survey_2021}. \textit{VITS} (Variational Inference with adversarial learning for end-to-end Text-to-Speech) \cite{kim_conditional_2021} is the first end-to-end model that achieves close to human quality regarding the naturalness of the synthesized speech. Its end-to-end approach also leads to improvements regarding synthesis speed. When trained on a multi-speaker dataset it also allows to switch between speaking styles (e.g. male or female voices) at inference time by selecting different speakers.
The most important aspects for VITS' positive results are a combination of different generative ML model approaches, namely VAEs (Variational Autoencoders) and GANs (Generative Adversarial Networks), as well as the newly proposed stochastic duration predictor to synthesize speech with diverse rhythms that helps to learn the one-to-many mapping from text to speech.
Simulating different emotions, cloning speakers' voices at inference time, or combining multiple languages in single models are active research topics at present \cite{tan_survey_2021}. There are multiple models openly available that provide different capabilities, cf.~\cite{kim_conditional_2021, casanova_yourtts_2022}. Additionally, open source libraries like \textit{coqui.ai}\footnote{https://github.com/coqui-ai/STT} allow to easily use and try out different models.

\paragraph{Chat} Since the introduction of Transformers \cite{vaswani_attention_2017}, language models with impressive capabilities were quickly developed. Especially GPT (Generative Pre-trained Transformer) models show, that scaling them up to more parameters and training them with more data makes them suitable to work on different natural language processing (NLP) tasks. The initial GPT \cite{radford_improving_2018} serves as pre-trained model that requires fine-tuning to work on tasks other than next token prediction. GPT-2 \cite{radford_language_2019} was already shown to generalize to other tasks more (especially to reading comprehension) or less (e.g., to summarization or translation) successfully, in a zero-shot fashion, i.e.~without requiring additional training.

DialoGPT \cite{zhang_dialogpt_2020} exploits GPT-2's architecture and is trained on conversation-like texts extracted from Reddit comments and thus it can be used as an open domain chatbot. Besides DialoGPT there are other Large Language Models (LLMs) specifically trained as open domain chatbots, e.g. BlenderBot \cite{roller_recipes_2020, xu_beyond_2021, shuster_blenderbot_2022}, which in contrast to GPT employs an encoder-decoder architecture. It outperforms DialoGPT in multi-turn conversations and in its development of three different versions, important aspects are considered regarding safety, long-term memory, sticking to a persona and integrating information from external sources.

Further scaling up GPT shows that GPT-3 \cite{brown_language_2020} is applicable to many different NLP tasks without fine-tuning but instead by providing it textual instructions on what to do or additional one or multiple examples, which is referred to as zero\nobreakdash-, one- or few-shot learning. Applying this approach GPT\nobreakdash-3 achieved performances on various NLP benchmark tasks nearly matching state of the art fine tuned systems. Also GPT-3's text generation capabilities reached a level at which human evaluators have difficulties to distinguish generated texts from human written texts: Human judges could identify news articles generated in a few-shot setting with an accuracy of 52\%, where 50\% is chance level performance \cite{brown_language_2020}. Given an appropriate prompt GPT-3 can be used as an open domain chatbot as well. 

The follow up GPT model was not simply scaled up, but instead fine-tuned with human feedback to better align to users' intents. The new model called InstructGPT \cite{ouyang_training_2022} was first fine-tuned in a supervised manner and then further fine-tuned using reinforcement learning from human feedback (RLHF). This feedback was collected using human labelers asked to rank different model outputs. 
In contrast to InstructGPT, ChatGPT was fine-tuned with differently collected data, to be usable in a dialogue format, that allows for follow up questions.

While GPT-2, as well as DialoGPT and all three versions of BlenderBot, models are openly accessible, this is not the case for the latest and most powerful models anymore. For example, ChatGPT is only accessible after registration via a Web-UI or a REST-API and a payed service after some amount of free usage is consumed. However there are promising open source solutions like Open Assistant\footnote{https://github.com/LAION-AI/Open-Assistant} following up.

\paragraph{Lip-Sync} There are multiple ML models that predict corresponding facial expressions, especially lip-movements for an input speech signal. While they are most often made for computer animation, they differ in the representation of facial expressions they generate as output: \textit{VisemeNet} \cite{zhou_visemenet_2018} predicts visemes (visually different expressions of the mouth during speech), in other works facial landmarks are predicted \cite{eskimez_generating_2018,eskimez_noise-resilient_2020} and \textit{FaceFormer}\cite{fan_faceformer_2022} and others\cite{karras_audio-driven_2017, cudeiro_capture_2019,richard_meshtalk_2022} predict a 3D-mesh of a whole virtual human head. In \cite{heisler_making_2023} we explored how predicted visemes and face meshes can be applied to animate an android robot head.

\section{Implementation}\label{implementation}

In this section first the current setup of the android robot head and its capabilities are described. Next some details on the development process are provided.

\subsection{Current state}
\begin{figure*}[htbp]
\centerline{\includegraphics[width=0.95\textwidth]{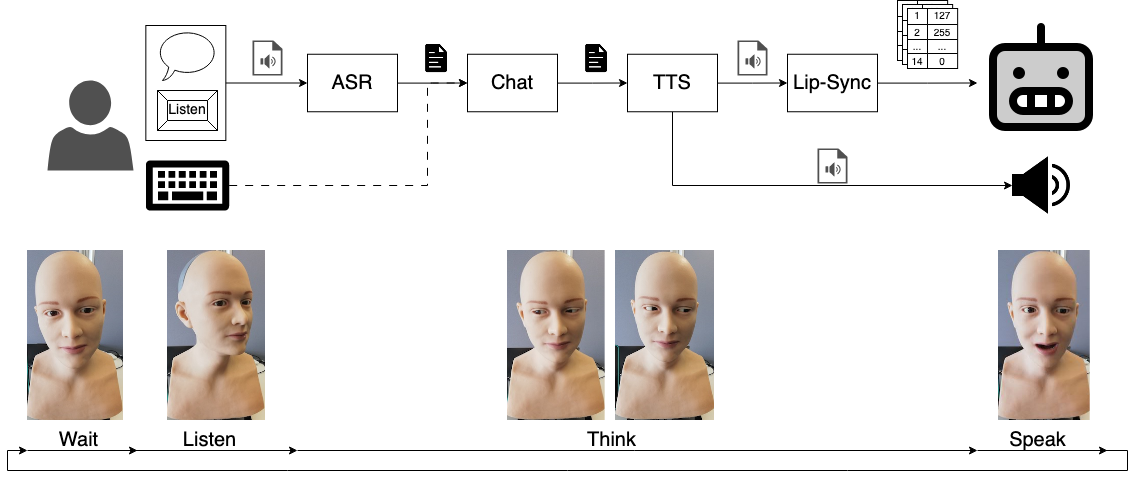}}
\caption{Overview of the current implementation. Top: basic pipeline from user input to response spoken by the android robot head. Bottom: animation phases according to the current timestep of the pipeline.}
\label{overview}
\end{figure*}

\begin{figure}[htp]
     \centering
     \begin{subfigure}[b]{0.45\textwidth}
         \centering
         \includegraphics[width=\textwidth]{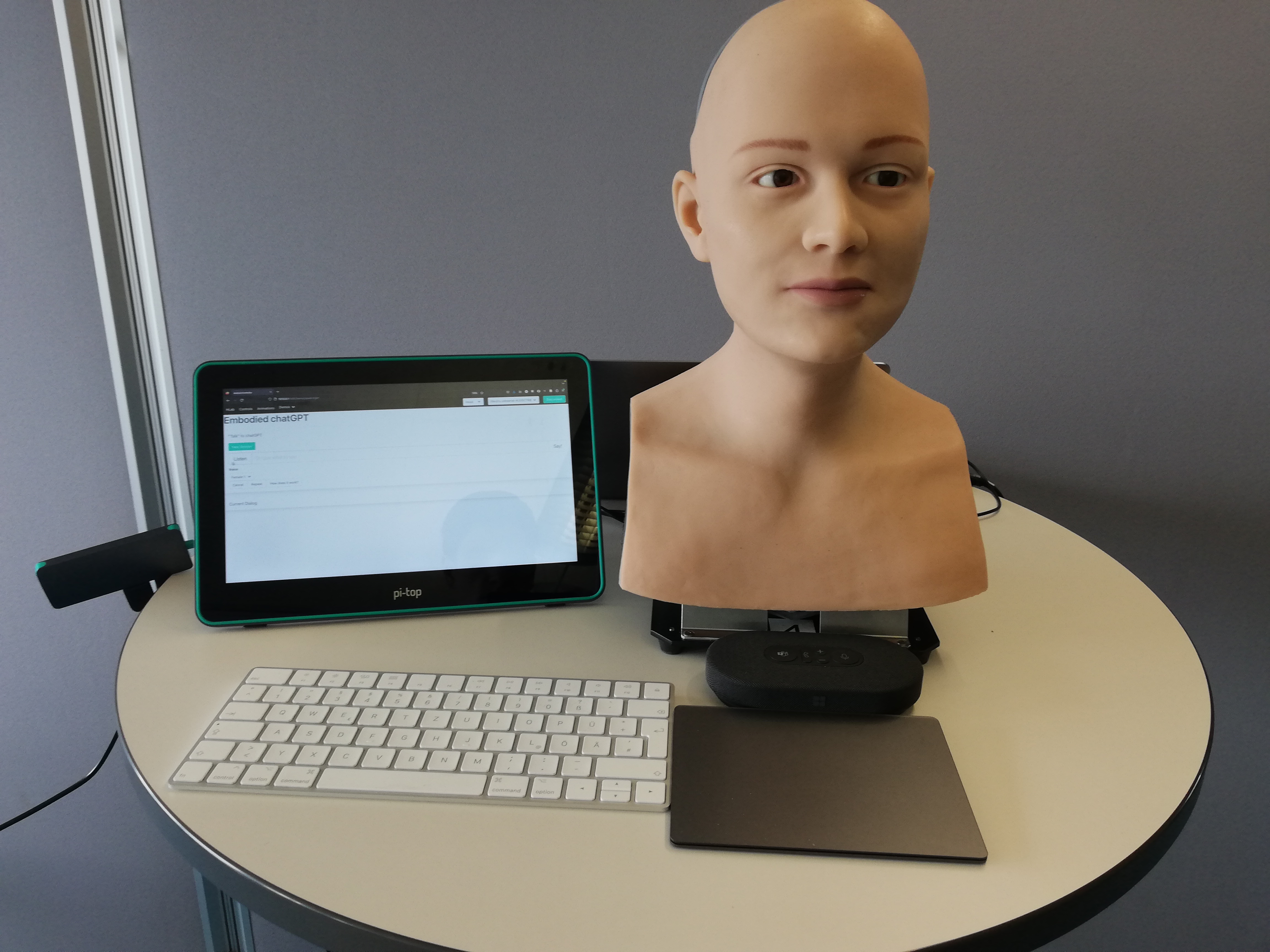}
         \caption{Setup as presented to users.}
         \label{setup}
     \end{subfigure}
     \hfill
     \begin{subfigure}[b]{0.45\textwidth}
         \centering
         \includegraphics[width=\textwidth]{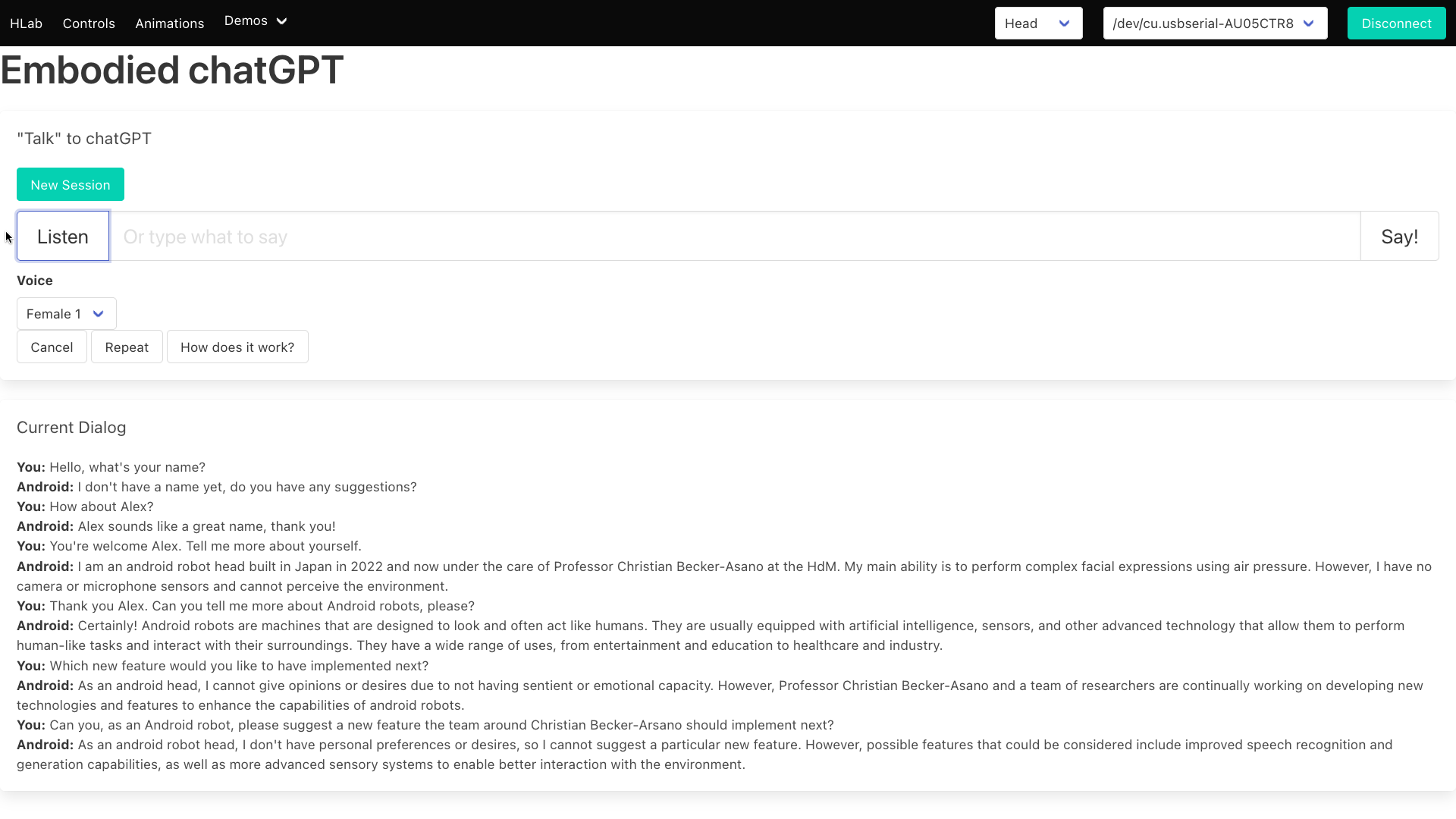}
         \caption{Exemplary screenshot of GUI.}
         \label{gui}
     \end{subfigure}
     \hfill
    \caption{Current setup of embodied conversational agent application.}
    \label{complete-setup}
\end{figure}

At the time of writing the android robot is able to converse in spoken natural language. The system is able to transcribe spoken language into text, generate a textual response, generate a speech signal articulating this response and display according facial expressions synchronous to its speech audio being output via an external speaker. This processing pipeline is shown in Fig.~\ref{overview} (top). The system is currently not able to detect, if it is talked to automatically, but instead relies on a push-to-talk button. Speech as input is optional and, if preferred by the user or required in very noisy environments, can be replaced by text input. The conversational capabilities of the system are combined with manually predefined animations of the android head. As shown in the lower part of Fig.~\ref{overview} such animations are automatically scheduled depending on the state of a conversation. Waiting animations are simple looping routines like random saccades and slight head movements and can be selected manually. Other such animations comprise randomized blinking movements and keep running during a conversation turn. Fig.~\ref{complete-setup} provides an impression of how the application is presented to a user.

Animations are defined in a custom data structure and stored as JSON files. Each animation defines a set of key frames, where each frame contains values for the 14 actuators of the robot head and a frame number. Values can be undefined and will then be replaced with values from a different active animation or the previously used values will be resent. Besides key frames each animation contains additional information further describing the animation. E.g., animations can contain absolute or relative values, where absolute means the actuator should move to the specified value of a that key frame, while relative means the actuators value should be increased or decreased from its current value accordingly. Also, it is possible to define, if and how values between consecutive key frames should be interpolated (though how is currently limited to linear). Another option to define is how many times an animation should run after it has been activated. Common values are once and looping until explicitly stopped. For animations to be executed multiple times, pauses can be specified as exact duration or as range to sample a random duration from. Additionally a priority can be defined for each animation to ensure that more important animations like lip-movements during speech are not compromised by any other less important animation possibly looping in the background. Finally some meta data like a name and description can be provided to be displayed in a GUI.

Besides manually starting and stopping animations, the current GUI implementation also allows to manipulate the values of each actuator separately using sliders. For the conversation application the GUI provides the before mentioned push-to-talk button and text input field, as well as possibilities to interrupt the current utterance or repeat the last one, to select a specific voice from the speech synthesizer and to start a new conversation by resetting the session of the chat module. Additionally it displays a conversation's turns so far (see Fig.~\ref{gui}).


The GUI is implemented as frontend of a web application written in \verb|python|. \verb|python| was chosen as programming language for multiple reasons: First, its low entry hurdle allows the educational use of the robotic head in lectures and semester projects with practical programming assignments even in lower bachelor semesters. Second, \verb|python| is quite common in robotics (e.g. besides \verb|C++|, ROS\cite{macenski_robot_2022} also provides \verb|python| APIs) and \verb|python| is the most used programming language for ML applications and research. Thus it is well suited to be used for our application, where both aspects are important. There are also multiple reasons to design the application as a web application: First, this is again quite simple because \verb|python| provides many frameworks to ease the development of web applications, e.g. flask\footnote{https://flask.palletsprojects.com/en/2.3.x/}, django\footnote{https://www.djangoproject.com/}, sanic\footnote{https://sanic.dev/en/} and FastAPI\footnote{https://fastapi.tiangolo.com/}. Here flask was chosen due to prior experience of the authors. Second the REST APIs provided by the backend can be reused for other applications, e.g. it is planned to make the robot head play chess without providing a GUI. A program handling the chess game can use the REST endpoints to schedule animations as needed. Third, it is accessible via (internal) network, thus a stationary computer can host it and make it easily accessible, without having to establish a cable connection to the head. Finally, it is also easy to integrate services running on remote servers. 
In case of the here described application, such services are the ML models, that run on self-hosted as well as external GPU clusters. Their integration into the application via REST API calls makes them easily replaceable.

For the different tasks described in Section~\ref{ml_models} and depicted in Fig.~\ref{overview}, the following ML models are currently in use: 

For \textbf{ASR} the current state-of-the art, \textit{Whisper}, is used. The open-source implementation from HuggingFace\footnote{https://huggingface.co/openai/whisper-large} runs on a self-hosted GPU cluster. For access a REST API endpoint is implemented using FastAPI. The cluster's capabilities are sufficient to run the \verb|large-v2| model, which has the most parameters and, thus, the best accuracy, with reasonable inference times.

To generate a textual answer \textit{chatGPT} is used as the \textbf{chat} component. The \verb|gpt-3.5-turbo| is integrated using the OpenAI Python Library\footnote{https://github.com/openai/openai-python}. The following prompt is sent in the \verb|system| role once, at the beginning of a new conversation: 

\textit{``You are a friendly android robot head. You are at a ChatGPT-related event at the Stuttgart Media University (HdM). You were built in Japan in 2022 and Professor Christian Becker-Asano of the HdM is now responsible for you and performs research with you. You represent the android robots of the HdM. There are five android heads including you, and one android with a complete body. You have the ability to do complex facial expressions using air pressure. You have no camera and microphone sensors and cannot perceive the environment. You can talk by using an external speaker. You do not have a name yet, but you are open for suggestions. Keep your answers short by using a maximum of three sentences to respond. Generate plain text output only, no code or other formats. Only respond in English.''} 

As the second sentence suggests the conversational application of the android robot head was first presented at a public event about chatGPT\footnote{https://ai.hdm-stuttgart.de/news/2023/event-resume-chatgpt-nur-ein-wenig-mathematik/}, which was also the main reason to use chatGPT over one of its open source alternatives. The prompt was not engineered\cite{weng_prompt_2023} but designed in a trial-and-error approach until suitable results were generated. 

For \textbf{TTS} coqui's implementation of \textit{VITS} is being used. Their \verb|tts-server| runs on our self-hosted GPU cluster and is called by the application via its REST API. Since the model provided by coqui is trained on the multi-speaker dataset VCTK\cite{yamagishi_cstr_2019}, 109 different English speakers are available at inference time. Two female and one male speakers with subjectively good quality are manually pre-selected and provided to choose from in the application's GUI. 

Finally, a \textit{FaceFormer} model wrapped in a FastAPI web application runs on our self-hosted GPU cluster, to automatically \textbf{lip-sync} the synthesized speech. Instead of rendering the generated sequence of points in 3D space, we map some manually defined distances between points onto actuator movements of the android robot head, cf.~\cite{heisler_making_2023} for details.

\subsection{Iterative development approach}

To achieve the current state of the implementation an iterative approach was employed, which is described in this subsection. While frameworks like Scrum\cite{schwaber_scrum_2020} define pretty clear guidelines on how to develop software in an agile and iterative way, practitioners often recommend to adapt the development process itself to the developer team's needs in an iterative fashion. Some aspects we found helpful during the development so far are to try out different approaches in a trial and error fashion but refactor afterwards to keep a clean code base, as well as to iteratively obtain feedback and adjust the most important next goals, which requires to create small but working increments.

Basically the following goals or milestones were reached during the development process so far:

\begin{enumerate}
\item Basic library
\item Multiple GUIs with sliders
\item Different Animations
\item Separate task oriented projects, e.g. lip-sync
\item Refactoring and integration
\item Speech synthesis (TTS)
\item Dialog (chatGPT)
\item Speech Recognition (ASR)
\end{enumerate}

First a basic library was implemented to establish a connection to the robot head and send values, which implies calculating CRC hash values. Using this basic library different GUIs were developed, that first of all allowed to control single actuators using sliders. Two types of GUIs were implemented in parallel: one web application and one pyQt\footnote{https://riverbankcomputing.com/software/pyqt/} based GUI. Next some basic animations were added to both of the GUIs. The definitions and implementations of the animations differed between the two types of GUIs: one focused on the interpolation between keyframes to easily define more complex animations like yawning, while the other concentrated on scheduling and combining multiple simple animations like blinking and saccades. Then different projects were implemented in parallel, each aiming at a specific task: gazing at people\footnote{https://ai.hdm-stuttgart.de/news/2023/gesichtstracking-mit-android-kopf/}, lip-syncing speech\cite{heisler_making_2023}, mimicking facial expressions and displaying emotional facial expressions.
These projects were built independently based on different GUI types, most of them as a student semester project. With the experience gained from these projects and newly identified requirements, like access via local network, it was decided to continue the web app and stop developing the other GUI. This required refactoring the current web app to integrate an animation schema combining the features of both independently developed solutions as well as the lip-synching capabilities.

Having a setup thus far in place, further capabilities were added mainly for demonstration purposes by plugging in different ML models. Implementing this was done in an order that allowed to have a version functioning and assessable on its own after each model added. Fist the TTS model was added to enable the robot head to speak arbitrary utterances, given as input text. Then a connection to ChatGPT was integrated for demonstration at the aforementioned public event. We used this event to collect feedback and identified two new features that were ask the most for: Speech input and multilingualism. By now speech input was added as final feature of the current implementation using \textit{Whisper}. 

\section{Evaluation and Outlook}\label{evaluation}

To evaluate the current setup, three lab members not working with any android robots themselves, were asked to have a conversation with the android robot head. Afterwards a semi-structured interview was conducted to find out two things: First, is the current implementation perceived as intended? Second, which aspect should be improved next?

Since the first question targets the perception, not the implementation details, its goal was to find out if the four phases of a conversation turn, as shown in the bottom part of Fig.~\ref{overview}, are recognized as such. While \textit{speak} and \textit{think} were clear to all of the participants, \textit{listen} and \textit{wait} (sometimes called \textit{idle} by interviewees) were less obvious at first, but also identifiable after a few conversational turns. 

In contrast to the feedback from the public presentation, none of the participants requested multilingualism as next new feature. Instead all participants see most potential in improving the existing animations. While the ``overacting'' during the \textit{think} phase is rated positive, the random eye and head movements during \textit{wait} and a sometimes appearing jerking behaviour during \textit{listen} are described as ``hectic'' and ``nervous'' and the fixed head and eye positions during \textit{listen} and \textit{speak} were perceived negative. Different solutions to improve the current animations are suggested: Two of the participants suggested to add an external camera to make the android robot head look at an interlocutor's eyes while speaking or at some areas with movements happening during the \textit{wait} phase. The other participant suggests that more subtle eye movements would already benefit both of the phases and some random nodding would help during listening. Also the participant that experienced the jerking bug recommends to fix this, of course.

Both adjusting animations and supporting multilingualism are desirable new features and manageable to implement with appropriate effort. To improve the gazing behaviour during animations the already implemented approach based on face detection with an external webcam and some heuristics, like moving the eyes first and turning the whole head a short time later, needs to be refactored and integrated into the current setup.
Since Whisper and ChatGPT already support multiple languages, only the TTS model needs to be replaced to enable multilingualism. Unfortunately the openly available \textit{yourTTS} model provided by coqui supports only English, Portuguese and French. Located in Germany ourselves we will need to train our own model before adding this feature. With datasets and code to train custom models publicly available the effort to implement this feature is estimated manageable.

Besides the features requested by users, we also plan to investigate into running the required ML models on an edge device. Though none of the users asked for faster \textit{thinking}, the shorter inference times might be beneficial, additionally we would like to become independent of a stable internet connection. Finally, although also not requested by any user we still plan to improve the lip-sync capabilities of the android robot head, as described in \cite{heisler_making_2023}.

Apart from fixing the jerking bug and some adjustments to make the current state work on our full-body android robot Andrea\footnote{https://ai.hdm-stuttgart.de/news/2022/mit-andrea-ist-man-ganz-vorne-dabei/}, a prioritisation of the possible next features was not carried out, yet.

\section{Conclusion}
The conceptualization and implementation of a very anthropomorphic robot head as an embodied conversational agent was presented. In doing so, it was highlighted how a combination of scripted animations and state-of-the-art ML models can achieve a convincing behavior in terms of timing and lip-sync animations. Most of the modules rely on open-source ML models, but unfortunately, the core component, ChatGPT, is closed source. This, together with general problems of privacy and legal risks, makes the current prototype not ready for commercial applications. Furthermore, it still remains a problem that the answers provided by large language models often tend to be ``too creative'' to rely on in any serious application. Our iterative development approach, however, enables us to test and compare different ML models systematically to evaluate the power of embodied AI for future applications.

\bibliographystyle{IEEEtran}
\bibliography{references}

\begin{thebibliography}{10}
\providecommand{\url}[1]{#1}
\csname url@samestyle\endcsname
\providecommand{\newblock}{\relax}
\providecommand{\bibinfo}[2]{#2}
\providecommand{\BIBentrySTDinterwordspacing}{\spaceskip=0pt\relax}
\providecommand{\BIBentryALTinterwordstretchfactor}{4}
\providecommand{\BIBentryALTinterwordspacing}{\spaceskip=\fontdimen2\font plus
\BIBentryALTinterwordstretchfactor\fontdimen3\font minus
  \fontdimen4\font\relax}
\providecommand{\BIBforeignlanguage}[2]{{%
\expandafter\ifx\csname l@#1\endcsname\relax
\typeout{** WARNING: IEEEtran.bst: No hyphenation pattern has been}%
\typeout{** loaded for the language `#1'. Using the pattern for}%
\typeout{** the default language instead.}%
\else
\language=\csname l@#1\endcsname
\fi
#2}}
\providecommand{\BIBdecl}{\relax}
\BIBdecl

\bibitem{kawahara_intelligent_2021}
\BIBentryALTinterwordspacing
T.~Kawahara, K.~Inoue, and D.~Lala, ``\BIBforeignlanguage{en}{Intelligent
  {Conversational} {Android} {ERICA} {Applied} to {Attentive} {Listening} and
  {Job} {Interview}},'' May 2021, number: arXiv:2105.00403 arXiv:2105.00403
  [cs]. [Online]. Available: \url{http://arxiv.org/abs/2105.00403}
\BIBentrySTDinterwordspacing

\bibitem{carros_not_2022}
F.~Carros, B.~B{\"u}rvenich, R.~Browne, Y.~Matsumoto, G.~Trovato, M.~Manavi,
  K.~Homma, T.~Ogawa, R.~Wieching, and V.~Wulf, ``Not that uncanny after all?
  an ethnographic study on android robots perception of older adults
  in germany and japan,'' in \emph{Social Robotics}, F.~Cavallo, J.-J.
  Cabibihan, L.~Fiorini, A.~Sorrentino, H.~He, X.~Liu, Y.~Matsumoto, and S.~S.
  Ge, Eds.\hskip 1em plus 0.5em minus 0.4em\relax Cham: Springer Nature
  Switzerland, 2022, pp. 574--586.

\bibitem{sato_android_2022}
\BIBentryALTinterwordspacing
W.~Sato, S.~Namba, D.~Yang, S.~Nishida, C.~Ishi, and T.~Minato,
  ``\BIBforeignlanguage{en}{An {Android} for {Emotional} {Interaction}:
  {Spatiotemporal} {Validation} of {Its} {Facial} {Expressions}},''
  \emph{\BIBforeignlanguage{en}{Front. Psychol.}}, vol.~12, p. 800657, Feb.
  2022. [Online]. Available:
  \url{https://www.frontiersin.org/articles/10.3389/fpsyg.2021.800657/full}
\BIBentrySTDinterwordspacing

\bibitem{glas_erica_2016}
\BIBentryALTinterwordspacing
D.~F. Glas, T.~Minato, C.~T. Ishi, T.~Kawahara, and H.~Ishiguro,
  ``\BIBforeignlanguage{en}{{ERICA}: {The} {ERATO} {Intelligent}
  {Conversational} {Android}},'' in \emph{\BIBforeignlanguage{en}{2016 25th
  {IEEE} {International} {Symposium} on {Robot} and {Human} {Interactive}
  {Communication} ({RO}-{MAN})}}.\hskip 1em plus 0.5em minus 0.4em\relax New
  York, NY, USA: IEEE, Aug. 2016, pp. 22--29. [Online]. Available:
  \url{http://ieeexplore.ieee.org/document/7745086/}
\BIBentrySTDinterwordspacing

\bibitem{minato_study_2022}
\BIBentryALTinterwordspacing
T.~Minato, K.~Sakai, T.~Uchida, and H.~Ishiguro, ``A study of interactive robot
  architecture through the practical implementation of conversational
  android,'' \emph{Frontiers in Robotics and AI}, vol.~9, 2022. [Online].
  Available:
  \url{https://www.frontiersin.org/articles/10.3389/frobt.2022.905030}
\BIBentrySTDinterwordspacing

\bibitem{cassell2001embodied}
J.~Cassell, ``Embodied conversational agents: representation and intelligence
  in user interfaces,'' \emph{AI magazine}, vol.~22, no.~4, pp. 67--67, 2001.

\bibitem{heisler_making_2023}
M.~Heisler, S.~Kopp, and C.~Becker-Asano, ``Making an android robot head
  talk,'' in \emph{2023 {IEEE} {International} {Symposium} on {Robot} and
  {Human} {Interactive} {Communication} ({RO}-{MAN})}.\hskip 1em plus 0.5em
  minus 0.4em\relax IEEE, under review.

\bibitem{radford_robust_2022}
\BIBentryALTinterwordspacing
A.~Radford, J.~W. Kim, T.~Xu, G.~Brockman, C.~McLeavey, and I.~Sutskever,
  ``Robust {Speech} {Recognition} via {Large}-{Scale} {Weak} {Supervision},''
  Dec. 2022, arXiv:2212.04356 [cs, eess]. [Online]. Available:
  \url{http://arxiv.org/abs/2212.04356}
\BIBentrySTDinterwordspacing

\bibitem{tan_survey_2021}
\BIBentryALTinterwordspacing
X.~Tan, T.~Qin, F.~Soong, and T.-Y. Liu, ``\BIBforeignlanguage{en}{A {Survey}
  on {Neural} {Speech} {Synthesis}},'' Jul. 2021, number: arXiv:2106.15561
  arXiv:2106.15561 [cs, eess]. [Online]. Available:
  \url{http://arxiv.org/abs/2106.15561}
\BIBentrySTDinterwordspacing

\bibitem{kim_conditional_2021}
\BIBentryALTinterwordspacing
J.~Kim, J.~Kong, and J.~Son, ``Conditional variational autoencoder with
  adversarial learning for end-to-end text-to-speech,'' in \emph{Proceedings of
  the 38th International Conference on Machine Learning}, ser. Proceedings of
  Machine Learning Research, M.~Meila and T.~Zhang, Eds., vol. 139.\hskip 1em
  plus 0.5em minus 0.4em\relax PMLR, 18--24 Jul 2021, pp. 5530--5540. [Online].
  Available: \url{https://proceedings.mlr.press/v139/kim21f.html}
\BIBentrySTDinterwordspacing

\bibitem{casanova_yourtts_2022}
E.~Casanova, J.~Weber, C.~D. Shulby, A.~C. Junior, E.~G{\"o}lge, and M.~A.
  Ponti, ``Yourtts: Towards zero-shot multi-speaker tts and zero-shot voice
  conversion for everyone,'' in \emph{International Conference on Machine
  Learning}.\hskip 1em plus 0.5em minus 0.4em\relax PMLR, 2022, pp. 2709--2720.

\bibitem{vaswani_attention_2017}
A.~Vaswani, N.~Shazeer, N.~Parmar, J.~Uszkoreit, L.~Jones, A.~N. Gomez,
  L.~Kaiser, and I.~Polosukhin, ``Attention is {All} you {Need},'' in
  \emph{Advances in {Neural} {Information} {Processing} {Systems}},
  vol.~30.\hskip 1em plus 0.5em minus 0.4em\relax Curran Associates, Inc.,
  2017.

\bibitem{radford_improving_2018}
A.~Radford, K.~Narasimhan, T.~Salimans, and I.~Sutskever,
  ``\BIBforeignlanguage{en}{Improving {Language} {Understanding} by
  {Generative} {Pre}-{Training}},'' 2018.

\bibitem{radford_language_2019}
A.~Radford, J.~Wu, R.~Child, D.~Luan, D.~Amodei, and I.~Sutskever,
  ``\BIBforeignlanguage{en}{Language {Models} are {Unsupervised} {Multitask}
  {Learners}},'' 2019.

\bibitem{zhang_dialogpt_2020}
Y.~Zhang, S.~Sun, M.~Galley, Y.-C. Chen, C.~Brockett, X.~Gao, J.~Gao, J.~Liu,
  and B.~Dolan, ``{DialoGPT}: {Large}-{Scale} {Generative} {Pre}-training for
  {Conversational} {Response} {Generation},'' in \emph{ACL, system
  demonstration}, 2020.

\bibitem{roller_recipes_2020}
\BIBentryALTinterwordspacing
S.~Roller, E.~Dinan, N.~Goyal, D.~Ju, M.~Williamson, Y.~Liu, J.~Xu, M.~Ott,
  K.~Shuster, E.~M. Smith, Y.-L. Boureau, and J.~Weston, ``Recipes for building
  an open-domain chatbot,'' Apr. 2020, number: arXiv:2004.13637
  arXiv:2004.13637 [cs]. [Online]. Available:
  \url{http://arxiv.org/abs/2004.13637}
\BIBentrySTDinterwordspacing

\bibitem{xu_beyond_2021}
\BIBentryALTinterwordspacing
J.~Xu, A.~Szlam, and J.~Weston, ``Beyond {Goldfish} {Memory}: {Long}-{Term}
  {Open}-{Domain} {Conversation},'' Jul. 2021, arXiv:2107.07567 [cs]. [Online].
  Available: \url{http://arxiv.org/abs/2107.07567}
\BIBentrySTDinterwordspacing

\bibitem{shuster_blenderbot_2022}
\BIBentryALTinterwordspacing
K.~Shuster, J.~Xu, M.~Komeili, D.~Ju, E.~M. Smith, S.~Roller, M.~Ung, M.~Chen,
  K.~Arora, J.~Lane, M.~Behrooz, W.~Ngan, S.~Poff, N.~Goyal, A.~Szlam, Y.-L.
  Boureau, M.~Kambadur, and J.~Weston, ``{BlenderBot} 3: a deployed
  conversational agent that continually learns to responsibly engage,'' Aug.
  2022, arXiv:2208.03188 [cs]. [Online]. Available:
  \url{http://arxiv.org/abs/2208.03188}
\BIBentrySTDinterwordspacing

\bibitem{brown_language_2020}
\BIBentryALTinterwordspacing
T.~B. Brown, B.~Mann, N.~Ryder, M.~Subbiah, J.~Kaplan, P.~Dhariwal,
  A.~Neelakantan, P.~Shyam, G.~Sastry, A.~Askell, S.~Agarwal, A.~Herbert-Voss,
  G.~Krueger, T.~Henighan, R.~Child, A.~Ramesh, D.~M. Ziegler, J.~Wu,
  C.~Winter, C.~Hesse, M.~Chen, E.~Sigler, M.~Litwin, S.~Gray, B.~Chess,
  J.~Clark, C.~Berner, S.~McCandlish, A.~Radford, I.~Sutskever, and D.~Amodei,
  ``Language {Models} are {Few}-{Shot} {Learners},'' Jul. 2020,
  arXiv:2005.14165 [cs]. [Online]. Available:
  \url{http://arxiv.org/abs/2005.14165}
\BIBentrySTDinterwordspacing

\bibitem{ouyang_training_2022}
\BIBentryALTinterwordspacing
L.~Ouyang, J.~Wu, X.~Jiang, D.~Almeida, C.~L. Wainwright, P.~Mishkin, C.~Zhang,
  S.~Agarwal, K.~Slama, A.~Ray, J.~Schulman, J.~Hilton, F.~Kelton, L.~Miller,
  M.~Simens, A.~Askell, P.~Welinder, P.~Christiano, J.~Leike, and R.~Lowe,
  ``Training language models to follow instructions with human feedback,'' Mar.
  2022, arXiv:2203.02155 [cs]. [Online]. Available:
  \url{http://arxiv.org/abs/2203.02155}
\BIBentrySTDinterwordspacing

\bibitem{zhou_visemenet_2018}
\BIBentryALTinterwordspacing
Y.~Zhou, Z.~Xu, C.~Landreth, E.~Kalogerakis, S.~Maji, and K.~Singh,
  ``{VisemeNet}: {Audio}-{Driven} {Animator}-{Centric} {Speech} {Animation},''
  May 2018, arXiv:1805.09488 [cs]. [Online]. Available:
  \url{http://arxiv.org/abs/1805.09488}
\BIBentrySTDinterwordspacing

\bibitem{eskimez_generating_2018}
\BIBentryALTinterwordspacing
S.~E. Eskimez, R.~K. Maddox, C.~Xu, and Z.~Duan, ``Generating {Talking} {Face}
  {Landmarks} from {Speech},'' Apr. 2018, arXiv:1803.09803 [cs] version: 2.
  [Online]. Available: \url{http://arxiv.org/abs/1803.09803}
\BIBentrySTDinterwordspacing

\bibitem{eskimez_noise-resilient_2020}
\BIBentryALTinterwordspacing
------, ``\BIBforeignlanguage{en}{Noise-{Resilient} {Training} {Method} for
  {Face} {Landmark} {Generation} {From} {Speech}},''
  \emph{\BIBforeignlanguage{en}{IEEE/ACM Trans. Audio Speech Lang. Process.}},
  vol.~28, pp. 27--38, 2020. [Online]. Available:
  \url{https://ieeexplore.ieee.org/document/8871109/}
\BIBentrySTDinterwordspacing

\bibitem{fan_faceformer_2022}
Y.~Fan, Z.~Lin, J.~Saito, W.~Wang, and T.~Komura, ``{FaceFormer}:
  {Speech}-{Driven} {3D} {Facial} {Animation} with {Transformers},'' in
  \emph{2022 {IEEE}/{CVF} {Conference} on {Computer} {Vision} and {Pattern}
  {Recognition} ({CVPR})}, Jun. 2022, pp. 18\,749--18\,758, iSSN: 2575-7075.

\bibitem{karras_audio-driven_2017}
\BIBentryALTinterwordspacing
T.~Karras, T.~Aila, S.~Laine, A.~Herva, and J.~Lehtinen,
  ``\BIBforeignlanguage{en}{Audio-driven facial animation by joint end-to-end
  learning of pose and emotion},'' \emph{\BIBforeignlanguage{en}{ACM Trans.
  Graph.}}, vol.~36, no.~4, pp. 1--12, Aug. 2017. [Online]. Available:
  \url{https://dl.acm.org/doi/10.1145/3072959.3073658}
\BIBentrySTDinterwordspacing

\bibitem{cudeiro_capture_2019}
\BIBentryALTinterwordspacing
D.~Cudeiro, T.~Bolkart, C.~Laidlaw, A.~Ranjan, and M.~J. Black,
  ``\BIBforeignlanguage{en}{Capture, {Learning}, and {Synthesis} of {3D}
  {Speaking} {Styles}},'' in \emph{\BIBforeignlanguage{en}{2019 {IEEE}/{CVF}
  {Conference} on {Computer} {Vision} and {Pattern} {Recognition}
  ({CVPR})}}.\hskip 1em plus 0.5em minus 0.4em\relax Long Beach, CA, USA: IEEE,
  Jun. 2019, pp. 10\,093--10\,103. [Online]. Available:
  \url{https://ieeexplore.ieee.org/document/8954000/}
\BIBentrySTDinterwordspacing

\bibitem{richard_meshtalk_2022}
\BIBentryALTinterwordspacing
A.~Richard, M.~Zollhoefer, Y.~Wen, F.~de~la Torre, and Y.~Sheikh, ``{MeshTalk}:
  {3D} {Face} {Animation} from {Speech} using {Cross}-{Modality}
  {Disentanglement},'' May 2022, arXiv:2104.08223 [cs]. [Online]. Available:
  \url{http://arxiv.org/abs/2104.08223}
\BIBentrySTDinterwordspacing

\bibitem{macenski_robot_2022}
\BIBentryALTinterwordspacing
S.~Macenski, T.~Foote, B.~Gerkey, C.~Lalancette, and W.~Woodall, ``Robot
  operating system 2: Design, architecture, and uses in the wild,''
  \emph{Science Robotics}, vol.~7, no.~66, p. eabm6074, 2022. [Online].
  Available: \url{https://www.science.org/doi/abs/10.1126/scirobotics.abm6074}
\BIBentrySTDinterwordspacing

\bibitem{weng_prompt_2023}
\BIBentryALTinterwordspacing
L.~Weng, ``Prompt engineering,'' \emph{lilianweng.github.io}, Mar 2023.
  [Online]. Available:
  \url{https://lilianweng.github.io/posts/2023-03-15-prompt-engineering/}
\BIBentrySTDinterwordspacing

\bibitem{yamagishi_cstr_2019}
J.~Yamagishi, C.~Veaux, and K.~MacDonald, ``Cstr vctk corpus: English
  multi-speaker corpus for cstr voice cloning toolkit,'' 2019.

\bibitem{schwaber_scrum_2020}
\BIBentryALTinterwordspacing
K.~Schwaber and J.~Sutherland, ``The scrum guide,'' 2020. [Online]. Available:
  \url{https://scrumguides.org/index.html}
\BIBentrySTDinterwordspacing

\end{thebibliography}

\end{document}